\pdfoutput=1
\documentclass[sigconf]{acmart}

\usepackage{amsmath,bm}
\usepackage{tikz,pgfplots}
\usepackage{multirow}
\usepackage{balance}

\copyrightyear{2021} 
\acmYear{2021} 
\setcopyright{rightsretained} 
\acmConference[CIKM '21]{Proceedings of the 30th ACM International Conference on Information and Knowledge Management}{November 1--5, 2021}{Virtual Event, QLD, Australia}
\acmBooktitle{Proceedings of the 30th ACM International Conference on Information and Knowledge Management (CIKM '21), November 1--5, 2021, Virtual Event, QLD, Australia}
\acmPrice{}
\acmISBN{978-1-4503-8446-9/21/11}
\acmDOI{10.1145/3459637.3482177}

\begin{document}
\fancyhead{}

\title{SeDyT: A General Framework for Multi-Step Event Forecasting via Sequence Modeling on Dynamic Entity Embeddings}

\author{Hongkuan Zhou}
\author{James Orme-Rogers}
\email{{hongkuaz,ormeroge}@usc.edu}
\affiliation{%
  \institution{University of Southern California}
  \city{Los Angeles}
  \state{California}
  \country{USA}
}

\author{Rajgopal Kannan}
\email{rajgopal.kannan.civ@mail.mil}
\affiliation{%
  \institution{US Army Research Lab}
  \city{Los Angeles}
  \state{California}
  \country{USA}
}

\author{Viktor Prasanna}
\email{prasanna@usc.edu}
\affiliation{%
  \institution{University of Southern California}
  \city{Los Angeles}
  \state{California}
  \country{USA}
}






\begin{abstract}
  Temporal Knowledge Graphs store events in the form of subjects, relations, objects, and timestamps which are often represented by dynamic heterogeneous graphs. Event forecasting is a critical and challenging task in Temporal Knowledge Graph reasoning that predicts the subject or object of an event in the future. To obtain temporal embeddings multi-step away in the future, existing methods learn generative models that capture the joint distribution of the observed events. To reduce the high computation costs, these methods rely on unrealistic assumptions of independence and approximations in training and inference. In this work, we propose SeDyT, a discriminative framework that performs sequence modeling on the dynamic entity embeddings to solve the multi-step event forecasting problem. SeDyT consists of two components: a Temporal Graph Neural Network that generates dynamic entity embeddings in the past and a sequence model that predicts the entity embeddings in the future. Compared with the generative models, SeDyT does not rely on any heuristic-based probability model and has low computation complexity in both training and inference. SeDyT is compatible with most Temporal Graph Neural Networks and sequence models. We also design an efficient training method that trains the two components in one gradient descent propagation. We evaluate the performance of SeDyT on five popular datasets. By combining temporal Graph Neural Network models and sequence models, SeDyT achieves an average of 2.4\% MRR improvement when not using the validation set and more than 10\% MRR improvement when using the validation set.
\end{abstract}




\maketitle

\section{Introduction}
Knowledge Graphs (KGs) store real-world facts in the form of events while Temporal Knowledge Graphs (TKGs) store events with time information. Event prediction is an important task in TKG reasoning that discovers missing facts in TKGs. Event forecasting is a more challenging and more impactful task that predicts the events multi-step (timestamp) away in the future and can be used in many real-world problems like economic crisis prediction \cite{beets2009global}, crime prediction \cite{10.1145/2755492.2755496}, and epidemic modeling \cite{petropoulos2020forecasting}. In a TKG, events are represented by quadruplets $(s,r,o,t)$ that indicate subjects, relations, objects, and timestamps. An event that is valid for a period of time is decomposed into multiple events with the same $s$, $r$, and $o$ at different timestamps $t$.
The multi-step event forecasting task is defined by predicting the events in the future ($t>T$) giving the event quadruplets happen in the past ($t\leq T$). 
As forecasting the future events with unseen subjects, objects, or relations is extremely hard, researchers have narrowed the space to identify the subjects, objects, or relations of future events given the rest of the elements in the ground truth quadruplets.
In this work, we follow the most popular setup and focus on predicting the subject or object of the events in the future by answering the queries $(?,r,o,t)$ and $(s,r,?,t)$, which is equivalent to a multi-class classification problem with the number of candidate classes equals to the total number of entities.

Recently, Graph Neural Networks (GNNs) have demonstrated strong expressive power and high versatility in graph representation learning. To learn temporal information as well as structural and contextual information, Temporal GNNs \cite{evolvegcn,tgat} generate dynamic embeddings by adding time encodings to the node attributes and applying Recurrent Neural Networks (RNNs) to regulate the embeddings or the weights.
In the event forecasting problem, the quadruplet representation of TKG seamlessly forms a dynamic heterogeneous graph where the entities (subjects and objects) become the nodes and relations become the heterogeneous edges. The timestamps are attached to their corresponding edges. However, it is hard to directly apply Temporal GNNs to forecast future events on the dynamic heterogeneous graph due to the lack of future graph structure information. Inference and reusing the future graph structure step-by-step amplify the prediction error in the temporal embeddings, which leads to low accuracy.
To address this issue, previous works \cite{renet,nlsm} used generative models to fit the events into pre-defined heuristic-based probability systems. The future entity embeddings multi-step away are generated by integrating the probability model over all the future timestamps.
Compared with discriminative models, generative models typically have high computation costs and favor high-quality training data. The performance also profoundly depends on the design of the probability system.
In this work, we propose SeDyT, a {\it discriminative framework} that applies sequence modeling on the extracted dynamic embeddings to perform multi-step forecasting. The auxiliary sequence model overcomes the challenge of multi-step forecasting using Temporal GNN, allowing SeDyT to be trained in a simple discriminative manner. The main contributions of this work are:

\begin{itemize}
    \item We propose SeDyT, a highly flexible discriminative framework for event forecasting on TKGs which is compatible with most sequence models and Temporal GNNs. 
    \item We develop an efficient mini-batch training method that updates the parameters in the sequence model and the Temporal GNN in SeDyT in one gradient descent propagation. The co-training of the two components allows SeDyT to accurately learn the probability of each entity in the queries without any approximations or assumptions. 
    \item We evaluate the performance of SeDyT on five popular datasets under two different setups. Compared with the state-of-the-art baselines, SeDyT achieves the highest MRR on all five datasets.
\end{itemize}
\section{Background}

There are many works \cite{hyte,knowevolve,R-GCN,xu2020tero} on event prediction in TKGs with the task of predicting missing elements in the events. Recently, Re-NET \cite{renet} claimed to be the first to provide solution to the event forecasting task where only the ground truth in the training set is used at inference. Re-NET implemented a static GNN and several RNNs to capture the dynamic entity embeddings and mapped the embeddings to an auto-regressive probability model. Later, NLSM \cite{nlsm} proposed a variational inference technique to estimate the posterior distribution of an edge occurring between two specified nodes. In order to capture the joint distribution with low computation cost, these models have to stand on unpractical assumptions, such as the independence between two graph snapshots with a long time duration, two events happening at the same timestamp, and the statistical model parameters. To perform multi-step inference, these generative models need to integrate over the future steps which are also not computationally feasible. As a result, Re-NET samples in the probability space and NLSM approximates the posterior distribution using Evidence Lower Bound Objective (ELBO). By contrast, SeDyT does not need these assumptions and approximations in both training and inference. To guarantee accuracy, the probability of the target entity is directly computed by our discriminative model. Beside these generative models, CyGNet \cite{cygnet} tried to attach a copy module to predict the future from the history, which achieved state-of-the-art accuracy on some datasets.

\subsection{Temporal Heterogeneous Graph Neural Network}

GNN \cite{kipfgcn} and its variants \cite{mixhop,jumpingknowledge,gat,hgnn} generate node embeddings by iteratively gathering and aggregating neighbor information. The forward propagation in the $(k)^\text{th}$ GNN layer with attention mechanism is defined as

\begin{equation}
    \bm{h}_v^{(k)}=\sigma\left(\sum_{u\in\mathcal{N}(v)}\frac{\exp(\bm{Q}_v^{(k-1)}\cdot\bm{K}_u^{(k-1)})}{\sum_{u'\in\mathcal{N}(v)}\exp(\bm{Q}_v^{(k-1)}\cdot\bm{K}_{u'}^{(k-1)})}\bm{V}_u^{(k-1)}\right)
    \label{eq: gat}
\end{equation}
where $\bm{Q}_v^{(k-1)}=\bm{W}_q^{(k-1)}\bm{h}_v^{(k-1)}$ is the query, $\bm{K}_v^{(k-1)}=\bm{W}_k^{(k-1)}\bm{h}_v^{(k-1)}$ is the key, and $\bm{V}_v^{(k-1)}=\bm{W}_v^{(k-1)}\bm{h}_v^{(k-1)}$ is the value. $\sigma(\cdot)$ denotes the activation function. $\mathcal{N}(v)$ denotes the set of one-hop neighbors of node $v$.

To capture temporal information as well as structural and contextual information, recent works \cite{tgat,tgn,cawn} attached a functional time encoding that captures the time difference between the current time and the timestamp of the edge to traverse. Specifically, the neighbor messages $\bm{h}_u^{(k-1)}$ in Equation \ref{eq: gat} are concatenated with an additional functional time encoding

\begin{equation}
    \bm{\overline{h}}_u^{(k-1)}=\bm{h}_u^{(k-1)}||\cos(\bm{\omega}(t-t_{vu})+\bm{\phi})
    \label{eq: timeenc}
\end{equation}
where $t_{vu}$ is the timestamp of the edge from node $v$ to node $u$. $\bm{\omega}$ and $\bm{\phi}$ are two learnable vectors.

\subsection{Sequence Model}

Sequence models aim to predict the next or next several elements in a given sequence. Recurrent Neural Network (RNN) \cite{rnn} and its variants \cite{lstm, gru} capture the temporal dependency in the sequences by using the output in the previous cell as the input of the current cell. Transformers \cite{transformer} have proven to be powerful models in language sequences that rely purely on attention mechanisms. Convolutional Neural Networks (CNNs) and Multi-Layer Perceptrons (MLPs) can also be used in sequence modeling if the sequence length is fixed.
\section{Approach}

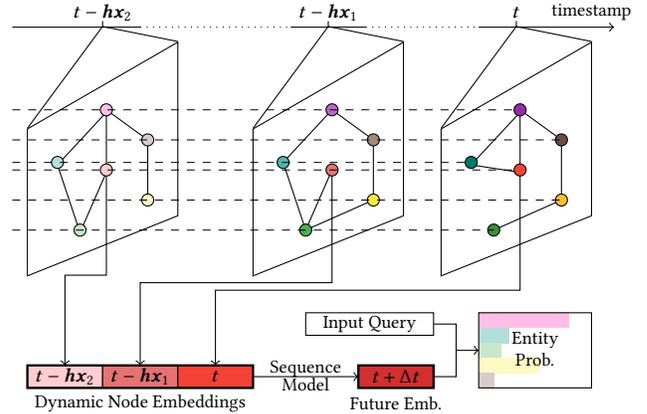
\begin{figure}[t!]
    \centering
    \begin{tikzpicture}[y=-1cm]

    \definecolor{r1}{RGB}{255,205,210}
    \definecolor{r2}{RGB}{229,115,115}
    \definecolor{r3}{RGB}{244,67,54}
    \definecolor{r4}{RGB}{211,47,47}
    
    \definecolor{p1}{RGB}{255,190,231}
    \definecolor{p2}{RGB}{186,104,200}
    \definecolor{p3}{RGB}{156,39,176}
    
    \definecolor{t1}{RGB}{178,223,219}
    \definecolor{t2}{RGB}{77,182,172}
    \definecolor{t3}{RGB}{0,121,107}
    
    \definecolor{g1}{RGB}{200,230,201}
    \definecolor{g2}{RGB}{76,175,80}
    \definecolor{g3}{RGB}{56,142,60}
    
    \definecolor{y1}{RGB}{255,249,196}
    \definecolor{y2}{RGB}{255,235,59}
    \definecolor{y3}{RGB}{251,192,45}

    \definecolor{b1}{RGB}{215,204,200}
    \definecolor{b2}{RGB}{161,136,127}
    \definecolor{b3}{RGB}{109,76,65}

    \node at (7.7,-0.2) {\footnotesize timestamp};
    \draw[-] (0,0) -- (1.7,0);
    \draw[dotted] (1.7,0) -- (3.7,0);
    \draw[-] (3.7,0) -- (4.7,0);
    \draw[dotted] (4.7,0) -- (6.2,0);
    \draw[->] (6.2,0) -- (8,0);
    \node at (1.2,-0.2) {\footnotesize $t-\bm{hx}_2$};
    \draw[-] (1.2,0) -- (1.2,-0.08);
    \node at (4.2,-0.2) {\footnotesize $t-\bm{hx}_1$};
    \draw[-] (4.2,0) -- (4.2,-0.08);
    \node at (6.7,-0.2) {\footnotesize $t$};
    \draw[-] (6.7,0) -- (6.7,-0.08);

    \coordinate (A) at (0.2,0.2);
    \draw[-] (A)++(0,1.1547) -- ++(2,-1.1547) -- ++(0,2.3094) -- ++(-2,0.8) -- ++(0,-1.9547);
    \draw[-] (A)++(0,1.1457) -- ++(1,-1.3547) -- ++(1,0.2);
    \draw[-] (A)++(1.05,1.7321) -- ++(0,-0.8) -- ++(-0.65,0.7) -- ++(0.3,0.9) -- ++(0.35,-0.8);
    \draw[-] (A)++(1.05,0.9) -- ++(0.55,0.4) -- ++(0,0.8);
    \draw[fill=r1] (A)++(1.05,1.7) circle (0.08);
    \draw[fill=p1] (A)++(1.05,0.9) circle (0.08);
    \draw[fill=t1] (A)++(0.4,1.6) circle (0.08);
    \draw[fill=g1] (A)++(0.7,2.5) circle (0.08);
    \draw[fill=y1] (A)++(1.6,2.1) circle (0.08);
    \draw[fill=b1] (A)++(1.6,1.3) circle (0.08);

    \coordinate (A) at (3.2,0.2);
    \draw[-] (A)++(0,1.1547) -- ++(2,-1.1547) -- ++(0,2.3094) -- ++(-2,0.8) -- ++(0,-1.9547);
    \draw[-] (A)++(0,1.1457) -- ++(1,-1.3547) -- ++(1,0.2);
    \draw[-] (A)++(1.05,0.9) -- ++(-0.65,0.7) -- ++(0.3,0.9) -- ++(0.35,-0.8);
    \draw[-] (A)++(1.05,0.9) -- ++(0.55,0.4) -- ++(0,0.8) -- ++(-0.9,0.4);
    \draw[fill=r2] (A)++(1.05,1.7) circle (0.08);
    \draw[fill=p2] (A)++(1.05,0.9) circle (0.08);
    \draw[fill=t2] (A)++(0.4,1.6) circle (0.08);
    \draw[fill=g2] (A)++(0.7,2.5) circle (0.08);
    \draw[fill=y2] (A)++(1.6,2.1) circle (0.08);
    \draw[fill=b2] (A)++(1.6,1.3) circle (0.08);

    \coordinate (A) at (5.7,0.2);
    \draw[-] (A)++(0,1.1547) -- ++(2,-1.1547) -- ++(0,2.3094) -- ++(-2,0.8) -- ++(0,-1.9547);
    \draw[-] (A)++(0,1.1457) -- ++(1,-1.3547) -- ++(1,0.2);
    \draw[-] (A)++(1.05,1.7321) -- ++(0,-0.8) -- ++(-0.65,0.7) -- ++(0.65,0.1);
    \draw[-] (A)++(1.05,0.9) -- ++(0.55,0.4) -- ++(0,0.8) -- ++(-0.9,0.4);
    \draw[fill=r3] (A)++(1.05,1.7) circle (0.08);
    \draw[fill=p3] (A)++(1.05,0.9) circle (0.08);
    \draw[fill=t3] (A)++(0.4,1.6) circle (0.08);
    \draw[fill=g3] (A)++(0.7,2.5) circle (0.08);
    \draw[fill=y3] (A)++(1.6,2.1) circle (0.08);
    \draw[fill=b3] (A)++(1.6,1.3) circle (0.08);

    \draw[dashed,line width=0.05mm] (0,1.1) -- (6.75,1.1);
    \draw[dashed,line width=0.05mm] (0,1.5) -- (7.3,1.5);
    \draw[dashed,line width=0.05mm] (0,1.8) -- (6.1,1.8);
    \draw[dashed,line width=0.05mm] (0,1.9) -- (6.75,1.9);
    \draw[dashed,line width=0.05mm] (0,2.3) -- (7.3,2.3);
    \draw[dashed,line width=0.05mm] (0,2.7) -- (6.4,2.7);

    \draw[fill=r1] (0.2,4.5) rectangle ++(1,0.3) node[pos=.5] {\footnotesize $t-\bm{hx}_2$};
    \draw[fill=r2] (1.2,4.5) rectangle ++(1,0.3) node[pos=.5] {\footnotesize $t-\bm{hx}_1$};
    \draw[fill=r3] (2.2,4.5) rectangle ++(1,0.3) node[pos=.5] {\footnotesize $t$};
    \draw[line width=0.3mm] (0.2,4.5) rectangle ++(3,0.3);
    \node at (1.7,5) {\footnotesize Dynamic Node Embeddings};
    
    \draw[->] (1.25,1.98) -- ++(0,1.32) -- ++(-0.55,0) -- ++(0,1.2);
    \draw[->] (4.25,1.98) -- ++(0,1.42) -- ++(-2.55,0) -- ++(0,1.1);
    \draw[->] (6.75,1.98) -- ++(0,1.52) -- ++(-4.05,0) -- ++(0,1);

    \draw[->] (3.2,4.65) -- ++(1.4,0);
    \node at (3.9,4.55) {\footnotesize Sequence};
    \node at (3.9,4.75) {\footnotesize Model};

    \draw[line width=0.3mm,fill=r4] (4.6,4.5) rectangle ++(1,0.3) node[pos=.5] {\footnotesize $t+\Delta t$};
    \node at (5.1,5) {\footnotesize Future Emb.};

    \draw (3.9,3.8) rectangle ++(1.7,0.3) node[pos=.5] {\footnotesize Input Query};
    \draw[->] (5.6,4.65) -- ++(0.3,0) -- ++(0,-0.35) -- ++(0.3,0);
    \draw[-] (5.6,3.95) -- ++(0.3,0) -- ++(0,0.35);
    \path[fill=p1] (6.2,3.8) rectangle ++(1.2,0.2);
    \path[fill=t1] (6.2,4) rectangle ++(0.4,0.2);
    \path[fill=g1] (6.2,4.2) rectangle ++(0.3,0.2);
    \path[fill=y1] (6.2,4.4) rectangle ++(0.8,0.2);
    \path[fill=b1] (6.2,4.6) rectangle ++(0.2,0.2);
    \draw (6.2,3.8) rectangle ++(1.5,1) node[pos=.5] {\footnotesize \begin{tabular}{c}Entity \\Prob.\end{tabular}};

\end{tikzpicture}
    \caption{Illustration of SeDyT forecasting an event $\Delta t$-steps away in the future with selected dynamic history embeddings $t-\bm{hx}$. The solid edges in each timestamp and the dashed lines connecting the entity at different timestamps represent the heterogeneous edges of the dynamic graph.}
    \label{fig:SeDyT}
\end{figure}

Figure \ref{fig:SeDyT} shows a simplified diagram of our SeDyT framework. To answer the query $(?,r,o,t+\Delta t)$ or $(s,r,?,t+\Delta t)$, SeDyT first computes the dynamic node embeddings of the input entity $o$ or $s$ at each of the history and current timestamps. Then, a sequence model generates the entity embeddings at the target timestamp $t+\Delta $. Finally, the classifier generates the probability of each candidate entity. We introduce these three processes in Section \ref{sec: dynamicnode}, Section \ref{sec: sequencemodel}, and Section \ref{sec: probabilitygeneration}, respectively. To efficiently train SeDyT in one gradient pass, we also design a novel training method, which is presented in Section \ref{sec: efficienttrain}.

\subsection{Dynamic Node Embedding Generation}
\label{sec: dynamicnode}

The dynamic node embeddings are generated by performing graph representation learning on the dynamic heterogeneous knowledge graph. In the knowledge graph, temporal heterogeneous edges represent events while static homogeneous nodes represent entities. For an event $(s,r,o,t)$, we add two edges to the knowledge graph: one $r$-typed edge from node $s$ to node $o$ with timestamp $t$ and one $\overline r$-typed edge from node $o$ to node $s$, which enables the entity nodes to gather information from both their corresponding objects and subjects. We use learnable static embeddings to serve as the input node attributes, which could also be replaced with entity features (if present in the dataset).

We adopt the multi-head attention \cite{gat} mechanism that performs multiple forward propagation in Equation \ref{eq: gat} with separate weight matrices $\bm W_q$, $\bm W_k$, and $\bm W_v$ in each attention head. Since the knowledge graph has heterogeneous edges, we use separate weight matrices for each edge type. Let $r\bm{h}^{(k)}_v$ be the aggregated message of node $v$ from its neighbors $\mathcal{N}_r(v)$ with $r$-typed edges. We use the mean aggregated message of all connected edge types $\Vert\mathcal{N}(v)\Vert$ as the hidden feature of node $v$.


To capture the temporal information in the edges, we replace $\bm{h}$ with $\overline{\bm{h}}$ in Equation \ref{eq: timeenc} except for the hidden features in the last layer. Thus, the generated embeddings contain the contextual information from node attributes as well as the structural and temporal information from temporal heterogeneous edges. Note that SeDyT is also compatible with traditional GNN+RNN methods \cite{renet,evolvegcn} to generate node embeddings.

\subsection{Sequence Modeling}
\label{sec: sequencemodel}

SeDyT uses the future entity embeddings to forecast events. Let $t$ be the current time. The future entity embedding $\bm{h}_v(t+\Delta t)$ of node $v$ at timestamp $t+\Delta t$ is computed by performing fixed-length sequence modeling on the sequence of the current and past entity embeddings $\left[\bm{h}_v(t-\bm{hx})\right]$ where $\bm{hx}$ denotes the selected history timestamps $\left[\bm{hx}_{|\bm{hx}|},\cdots,\bm{hx}_1,\bm{hx}_0=0\right]$. Sequence modeling is a well-studied problem, especially for fixed-length sequences. In this work, we focus on the following four sequence models.

\begin{itemize}
    \item SATT: Sequence modeling based on self-attention \cite{transformer} (encoder only). In each SATT layer, we compute the attention between each pair of elements in the entity embedding sequence. 
    \item CONV: Sequence modeling based on the convolutional operation. We stack the sequences vertically to obtain 2-D tensors and apply a $3\times3$ 2-D convolution kernel.
    \item MLP: Sequence modeling based on MLP. We concatenate the sequence horizontally and apply MLP to obtain the predicted future entity embedding.
    \item LSTM: Sequence modeling based on LSTM \cite{lstm}. We use a multi-to-one LSTM to predict the future entity embedding.
\end{itemize}

\subsection{Probability Generation}
\label{sec: probabilitygeneration}

To predict the subject or object of a future event, we take the future embedding of the corresponding object or subject and the learnable static relation embedding as input. We use MLP with residual connections to map the input to the probability of the candidate embeddings.

\subsection{Efficient Training}
\label{sec: efficienttrain}

To train SeDyT in a discriminative way, we set the step $\Delta t$ in the sequence model to be the time span of the validation and test set so that SeDyT can forecast all the events in the test set with the ground truth in the training set only. As a discriminative framework, SeDyT does not set a loss for the sequence model to predict the known history embeddings of the entity. Instead, we optimize the cross-entropy loss only between the generated probabilities and the ground truth entities for each event in the training set. However, directly performing gradient descent on this loss is infeasible because it must propagate on the graph structure which is different at the history timestamps $\bm{hx}$. 
To solve this problem, we create snapshots of window size $T$ by duplicating the nodes and adding a special type of edge between the same entity across the snapshots to retain connectivity as shown in Figure \ref{fig:SeDyT}. These self-connection edges are directed edges that only point from the past nodes to their future nodes, which avoids using the information from the future. This allows the Temporal GNN to compute forward and backward propagation at different timestamps simultaneously by computing the node embeddings of the same entity in different snapshots. 
\section{Experiments}

\begin{table*}[t!]
    \centering
    \caption{MRR, Hits@3, and Hit@10 results of event forecasting on the five datasets using the ground truth in the training set.}
    \setlength{\tabcolsep}{1.2mm}
    \begin{tabular}{r|ccc|ccc|ccc|ccc|ccc}
        \multirow{2}{*}{Method} & \multicolumn{3}{|c|}{GDELT} & \multicolumn{3}{|c|}{ICEWS14} & \multicolumn{3}{|c|}{ICEWS18} & \multicolumn{3}{|c|}{WIKI} & \multicolumn{3}{|c}{YAGO} \\
         & MRR & H@3 & H@10 & MRR & H@3 & H@10 & MRR & H@3 & H@10 & MRR & H@3 & H@10 & MRR & H@3 & H@10 \\
        \toprule
        R-GCN & 23.31 & 24.94 & 34.36 & 26.31 & 30.43 & 45.34 & 23.19 & 25.34 & 36.48 & 37.57 & 39.66 & 41.90 & 41.30 & 44.44 & 52.68 \\
        RotatE & 22.33 & 23.89 & 32.29 & 29.56 & 32.92 & 42.68 & 23.10 & 27.61 & 38.72 & 48.67 & 49.74 & 49.88 & 64.09 & 64.67 & 66.16\\
        HyTE & 6.37 & 6.72 & 18.63 & 11.48 & 13.04 & 22.51 & 7.31 & 7.50 & 14.95 & 43.02 & 45.12 & 49.49 & 23.16 & 45.74 & 51.94\\
        EvolveRGCN & 15.55 & 19.23 & 31.54 & 17.01 & 18.97 & 32.58 & 16.59 & 18.32 & 34.01 & 46.49 & 47.83 & 49.23 & 59.74 & 61.03 & 61.69\\
        R-GCRN+MLP & 37.29 & 41.08 & 51.88 & 36.77 & 40.15 & 52.33 & 35.12 & 38.26 & 50.49 & 47.71 & 48.14 & 49.66 & 53.89 & 56.06 & 61.19\\
        RE-NET & 40.42 & 43.40 & 53.70 & 45.71 & 49.06 & 59.12 & 42.93 & 45.47 & \textbf{55.80} & 51.97 & 52.07 & 53.91 & 65.16 & 65.63 & 68.08\\
        CyGNet & 50.22 & 53.26 & 57.42 & 48.41 & 52.23 & \textbf{59.82} & 45.82 & \textbf{48.62} & 55.14 & 45.23 & 50.81 & 52.12 & 64.42 & 65.02 & 67.59\\
        \midrule
        SeDyT-SATT & 54.96 & 54.79 & 58.30 & \textbf{52.76} & \textbf{52.89}  & 57.42 & 45.91 & 45.89 & 49.62 & 52.73 & 52.82 & 53.65 & 66.17 & 66.41 & 68.39 \\
        SeDyT-CONV & 54.86 & 54.68 & 58.14 & 52.72 & 52.86 & 57.43 & 45.91 & 45.86 & 49.54 & \textbf{52.90} & \textbf{52.96} & \textbf{54.00} & \textbf{66.88} & \textbf{67.05} & \textbf{68.73} \\
        SeDyT-MLP & \textbf{54.99} & \textbf{54.82} & \textbf{58.37} & 52.61 & 52.72 & 57.37 & 46.01 & 45.98 & 49.74 & 52.84 & 52.95 & 53.89 & 66.17 & 66.41 & 68.39 \\
        SeDyT-LSTM & 54.86 & 54.62 & 58.10 & 52.71 & 52.88 & 57.48 & \textbf{46.04} & 45.97 & 49.83 & 52.46 & 52.51 & 53.40 & 65.86 & 66.22 & 67.94 \\
    \end{tabular}
    \label{tab: main}
\end{table*}

\begin{table}[b!]
    \centering
    \caption{MRR, Hits@3, and Hit@10 results of event forecasting on WIKI and YAGO using the ground truth in the training and validation sets.}
    \setlength{\tabcolsep}{1.2mm}
    \begin{tabular}{r|ccc|ccc}
        \multirow{2}{*}{Method} & \multicolumn{3}{|c|}{WIKI} & \multicolumn{3}{|c}{YAGO} \\
         & MRR & H@3 & H@10 & MRR & H@3 & H@10 \\
        \toprule
        RE-NET & 53.57 & 54.10 & 55.72 & 66.80 & 67.23 & 69.77 \\
        NLSM & 56.70 & 57.80 & 61.10 & 69.40 & 71.25 & 73.90 \\
        \midrule
        SeDyT-SATT & 68.90 & 68.84 & 69.21 & 84.47 & 84.41 & 84.73\\
        SeDyT-CONV & 68.97 & 68.92 & 69.31 & \textbf{84.48} & \textbf{84.39} & 84.72\\
        SeDyT-MLP & \textbf{68.98} & \textbf{68.94} & \textbf{69.34} & \textbf{84.48} & 84.38 & \textbf{84.87}\\
        SeDyT-LSTM & 68.95 & 68.91 & 69.28 & 84.45 & 84.38 & 84.76\\
    \end{tabular}
    \label{tab: nlsm}
\end{table}

We evaluated the performance of SeDyT on five popular datasets: GDELT \cite{gdelt}, ICEWS14 \cite{knowevolve}, ICEWS18 \cite{icews18}, WIKI \cite{wiki}, and YAGO \cite{yago}. We follow the same training, validation, and test split used in the event forecasting works \cite{renet,nlsm,cygnet}. On ICEWS14, we ignore the timestamps of test events and set $\Delta t=1$ because the time duration in the test set is longer than the training set. We set 200 as the dimensions of the entity attributes and relation embeddings to be consistent with other works \cite{renet, cygnet}. In the Temporal GNN, we use a one-layer GNN described in Section \ref{sec: dynamicnode} to generate 200-dimensional temporal entity embeddings. We combine the heterogeneous edge types with low occurrences and use shared weights to perform message passing on these edges. The window size $T$ is set to be $4,3,3,1,1$ on the five datasets, respectively. We show the results of SeDyT with four different sequence models mentioned in Section \ref{sec: sequencemodel} with the history timestamps $\bm{hx}=[31,23,15,7,3,1,0]$. 
We adopt the copy module \cite{cygnet} which takes the temporal entity embeddings at the last timestamp $\bm{hx}_0=0$ as input and is disconnected in the gradient descent step. We use the ADAM optimizer \cite{adam} to train SeDyT till converge. We implement SeDyT using DGL \cite{dgl} and PyTorch. All the results are averages of three runs. Our code is available at \url{https://github.com/tedzhouhk/SeDyT}.

We first compare the performance of SeDyT with the static methods \cite{R-GCN,rotate} and the temporal methods \cite{hyte,evolvergcn,gcrn,renet,cygnet} under the setting of forecasting future events using the events in the training set only (i.e., $\Delta t$ equals to the time duration of the validation and test set). We modify the code of CyGNet so that both the subject and the object are predicted using a single model. Table \ref{tab: main} shows the filtered \cite{filter} MRR, Hit@3, and Hit@10 (in percentile) of SeDyT and the aforementioned baselines. The results of baselines except for CyGNet are taken from RE-NET \cite{renet}. SeDyT achieves the highest MRR on all datasets with an average improvement of 2.40\% over the state-of-the-art results. SeDyT also achieves the highest Hit@3 on all datasets and the highest Hit@10 on three datasets. The four variants of SeDyT achieves comparable performance, where SeDyT-CONV works best on the TKGs with the prolonged events (WIKI and YAGO). For the TKGs with point time events (GDELT, ICEWS14, and ICEWS18), SeDyT-MLP achieves best or close to best performance.

The setting of only using the ground truth in the training set at inference misspends the most up-to-date observed events that are in the validation set. To make use of the validation set at inference, we compare the performance of SeDyT with RE-NET \cite{renet} and NLSM \cite{nlsm} under the setting of forecasting future events using the events in the training and validation set (i.e., $\Delta t$ equals to the time duration of the test set). The results of the baselines are taken from NLSM \cite{nlsm}. With the additional ground truth in the validation set, SeDyT forecasts the future events with shorter $\Delta t$, which greatly boosts the performance. SeDyT-CONV and SeDyT-MLP achieve the best performance on these two datasets with more than 10\% improvement in MRR  compared with NLSM. 

\section{Conclusion and Future Work}

In this work, we proposed SeDyT, a discriminative framework for event forecasting on TKGs that is compatible with most Temporal GNNs and sequence models. We developed a snapshot-based TKG representation and an efficient training algorithm that trains the two components in SeDyT in one gradient pass. We evaluated the performance of four SeDyT variants on five datasets and showed that SeDyT achieved the highest MRR on all datasets.

In the future, we want to test the performance of SeDyT on large-scale datasets such as the GDELT dataset with multiple years instead of the one-month used in this work. 
As SeDyT is compatible with most Temporal GNNs, we would like to try more complex and powerful methods. 
In addition, under the current setting, SeDyT forecasts the future events with a fixed step $\Delta t$ where the ground truth events in the last several timestamps are not used when forecasting the events at the beginning several timestamps in the future. 
For example, when forecasting an event at timestamp $t+1$, SeDyT with fixed step $\Delta t$ ignores the ground truth events between $t-\Delta t+1$ to $t$. 
We plan to explore the sequence models that support adjustable sequence lengths like SATT and LSTM to support dynamic step prediction for SeDyT. 
Dynamic-step prediction also has the potential to improve the computation complexity when forecasting future events in a continuous-time interval.



\begin{acks}
This work is supported by the National Science Foundation (NSF) under grants OAC-1911229 and CNS-2009057 and in part by the Army Research Lab (ARL) under ARL-USC collaborative grant DIRA-ECI:DEC21-CI-037.
\end{acks}

\bibliographystyle{ACM-Reference-Format}
\balance
\bibliography{sample-base}


\end{document}